\documentclass[10pt,conference]{IEEEtran}
\IEEEoverridecommandlockouts
\usepackage{balance} 
\usepackage{float}
\usepackage{cite}
\usepackage{amsmath,amssymb,amsfonts}
\usepackage{textcomp}
\usepackage{xcolor}
\usepackage{color}
\usepackage{amsmath}
\usepackage[linesnumbered,ruled,vlined]{algorithm2e}
\usepackage{hyperref}
\usepackage{cleveref}
\hypersetup{hidelinks} 
\usepackage{multirow}
\usepackage{bm}
\usepackage[normalem]{ulem}

\usepackage{tikz}
\usepackage{pgfplots}
\usepackage{microtype}
\usepackage{graphicx}
\usepackage{booktabs} %
\usepackage{adjustbox}
\usepackage{xspace}
\usepackage{multirow}
\usepackage{thmtools, thm-restate} %

\ifCLASSOPTIONcompsoc
\else
\fi

\def\BibTeX{{\rm B\kern-.05em{\sc i\kern-.025em b}\kern-.08em
    T\kern-.1667em\lower.7ex\hbox{E}\kern-.125emX}}

\definecolor{transfer}{RGB}{199, 237, 204}  
\definecolor{meta}{RGB}{232, 209, 239}  
\definecolor{ewc}{RGB}{199, 228, 246}  

\begin{document}
\title{
EdgeFD: An Edge-Friendly Drift-Aware Fault Diagnosis System for Industrial IoT
}

\author{\IEEEauthorblockN{Jiao Chen\IEEEauthorrefmark{1}, Fengjian Mao\IEEEauthorrefmark{1}, Zuohong Lv\IEEEauthorrefmark{1}, and Jianhua Tang\IEEEauthorrefmark{1}\IEEEauthorrefmark{2} }
    \IEEEauthorblockA{\IEEEauthorrefmark{1} Shien-Ming Wu School of Intelligent Engineering, South China University of Technology, China}
    \IEEEauthorblockA{\IEEEauthorrefmark{2} Pazhou Lab, Guangzhou, China}
        \IEEEauthorblockA{\{202110190459,201930362330,202220159664\}@mail.scut.edu.cn, jtang4@e.ntu.edu.sg}

\thanks{This work was supported in part by the National Nature Science Foundation of China under Grant 62001168.
}
}
\maketitle

\begin{abstract}
Recent transfer learning (TL) approaches in industrial intelligent fault diagnosis (FD) mostly follow the ``pre-train and fine-tuning'' paradigm to address data drift, which emerges from variable working conditions. 
However, we find that this approach is prone to the phenomenon known as \textit{catastrophic forgetting}.
Furthermore, performing frequent models fine-tuning on the resource-constrained edge nodes can be computationally expensive and unnecessary, given the excellent transferability demonstrated by existing models.
In this work, we propose the Drift-Aware Weight Consolidation (DAWC), a method optimized for edge deployments, mitigating the challenges posed by frequent data drift in the industrial Internet of Things (IIoT).
DAWC efficiently manages multiple data drift scenarios, minimizing the need for constant model fine-tuning on edge devices, thereby conserving computational resources.
By detecting drift using classifier confidence and estimating parameter importance with the Fisher Information Matrix—a tool that measures parameter sensitivity in probabilistic models, we introduce a drift detection module and a continual learning module to gradually equip the FD model with powerful generalization capabilities.
Experimental results demonstrate that our proposed DAWC achieves superior performance compared to existing techniques while also ensuring compatibility with edge computing constraints.
Additionally, we have developed a comprehensive diagnosis and visualization platform.
The project webpage is \url{https://github.com/tenderzada/BearingEdge}.
\end{abstract}

\begin{IEEEkeywords}
Mechanical Fault Diagnosis, Drift Detection, Continual Learning, Industrial Internet of Things, Edge Computing
\end{IEEEkeywords}

\section{Introduction}
\label{section:introduction}

The integration of the Industrial Internet of Things (IIoT) and its key technologies (e.g., reconfigurable intelligent surface \cite{10200033}) into industrial operations has triggered a profound transformation in operational methodologies. IIoT harnesses interconnected sensors, devices, and machinery to capture real-time data, facilitating data-driven decision-making and predictive maintenance strategies \cite{10018896,zhao2022memory}.

In industry, bearings and harmonic reducers are pivotal mechanical components that support and transmit rotary motions within mechanical systems. However, bearing failures due to prolonged operation and load-bearing can result in equipment downtime and production disruptions, leading to significant economic losses and safety hazards. In response to these challenges, there exists a growing demand for intelligent fault diagnosis (FD), early warning, and visualization of bearing faults \cite{10102331,10113185,chen2022interpretable}.

In recent years, integrating edge computing \cite{7914660} into the IIoT framework have offered a promising avenue for enhancing intelligent diagnostic systems.
Edge computing involves processing and analyzing data at or near the data source, enabling real-time analysis and minimizing data transmission latency. Notably, the convergence of IIoT and edge computing renders the concept of ``source-end detection" in mechanical FD both practical and effective.

\begin{figure}[t]
    \centering
    \includegraphics[width=0.9\linewidth]{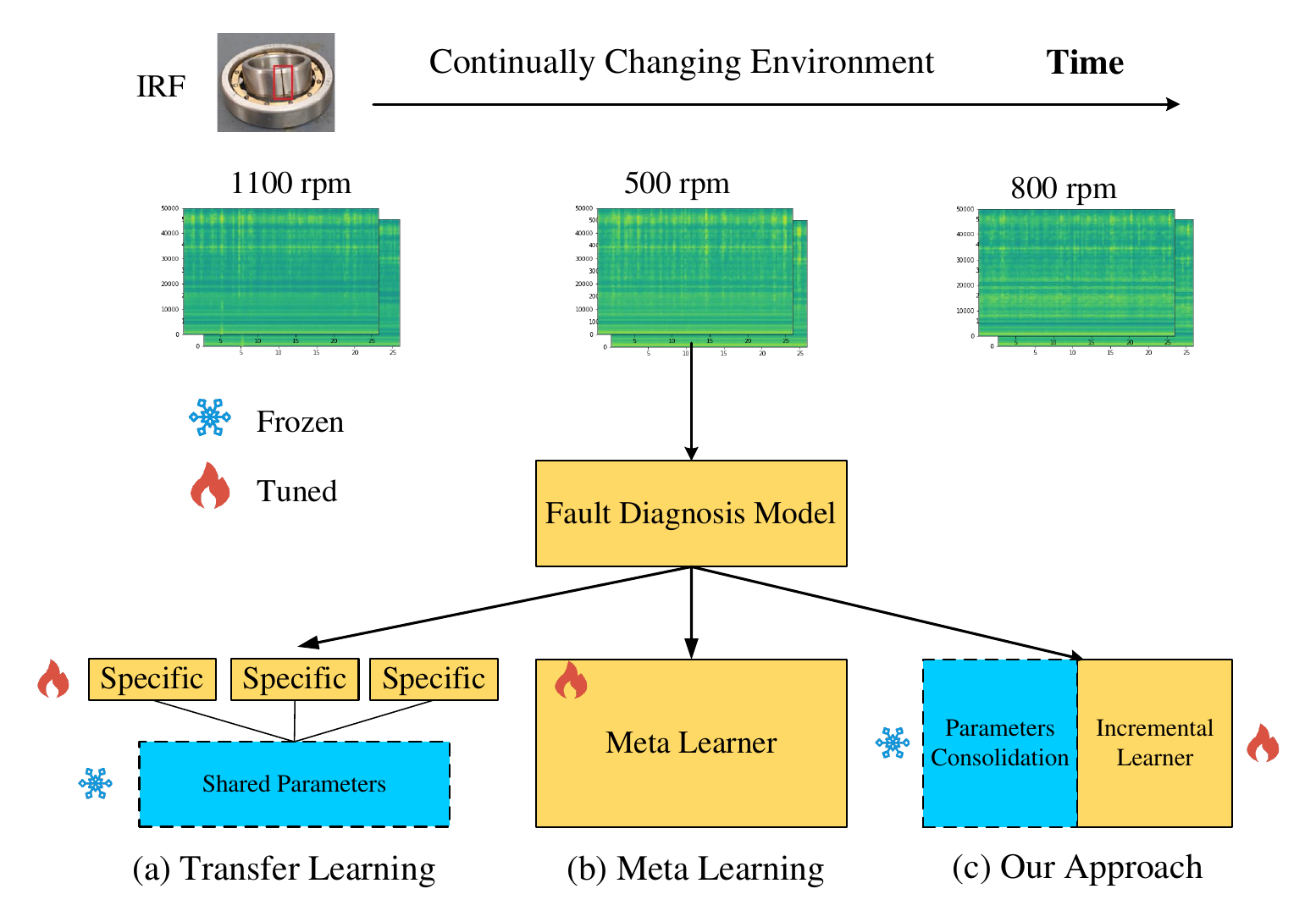}
    \caption{Comparative analysis of learning approaches in the context of evolving data distributions. The data distribution of a specific Inner Race Fault (IRF) changes over time due to varying rpm or loads. The objective is to create a model that effectively generalizes across this evolving data distribution.
    (a) Transfer Learning: Task-specific parameters are fine-tuned for each data drift.
    (b) Meta-Learning: The model is quickly adapted using a few samples from the new data distribution.
    (c) Our Approach: Important weights from past data distributions are consolidated while adapting to new tasks, ensuring a balance of specificity and generality.
    }
    \label{fig1}
\end{figure}

The potential of intelligent FD methods has been showcased in various studies, yet a formidable challenge persists, namely, data drift, which leads to a notable degradation in diagnosis accuracy \cite{bandyopadhyay2022intelligent}.
Data drift, encompassing challenges such as domain incremental learning (where new classes emerge over time), changes in prior probabilities, and covariate drift (changes in the distribution of input data), hampers the deployment of FD models on edge devices \cite{hurtado2023continual}.
In industrial settings, especially where data can be scarce or the environment rapidly evolves, this problem becomes particularly pronounced. 

To effectively address the issue of data drift in mechanical fault diagnosis, Transfer Learning (TL) \cite{chen2023transfer,chen2023aero} and Meta-Learning (ML) \cite{feng2022meta} are often employed. 
TL and ML address the data drift problem by facilitating the transfer of knowledge from previously learned tasks to new ones, which enhances model robustness and data efficiency—essential traits in resource-constrained or swiftly evolving industrial settings.
These methodologies promote better generalization across diverse fault scenarios. Moreover, ML excels in rapid adaptability to new fault conditions, pushing real-time fault diagnosis closer to realization.

\begin{figure}[t]
    \centering
    \includegraphics[width=0.9\linewidth]{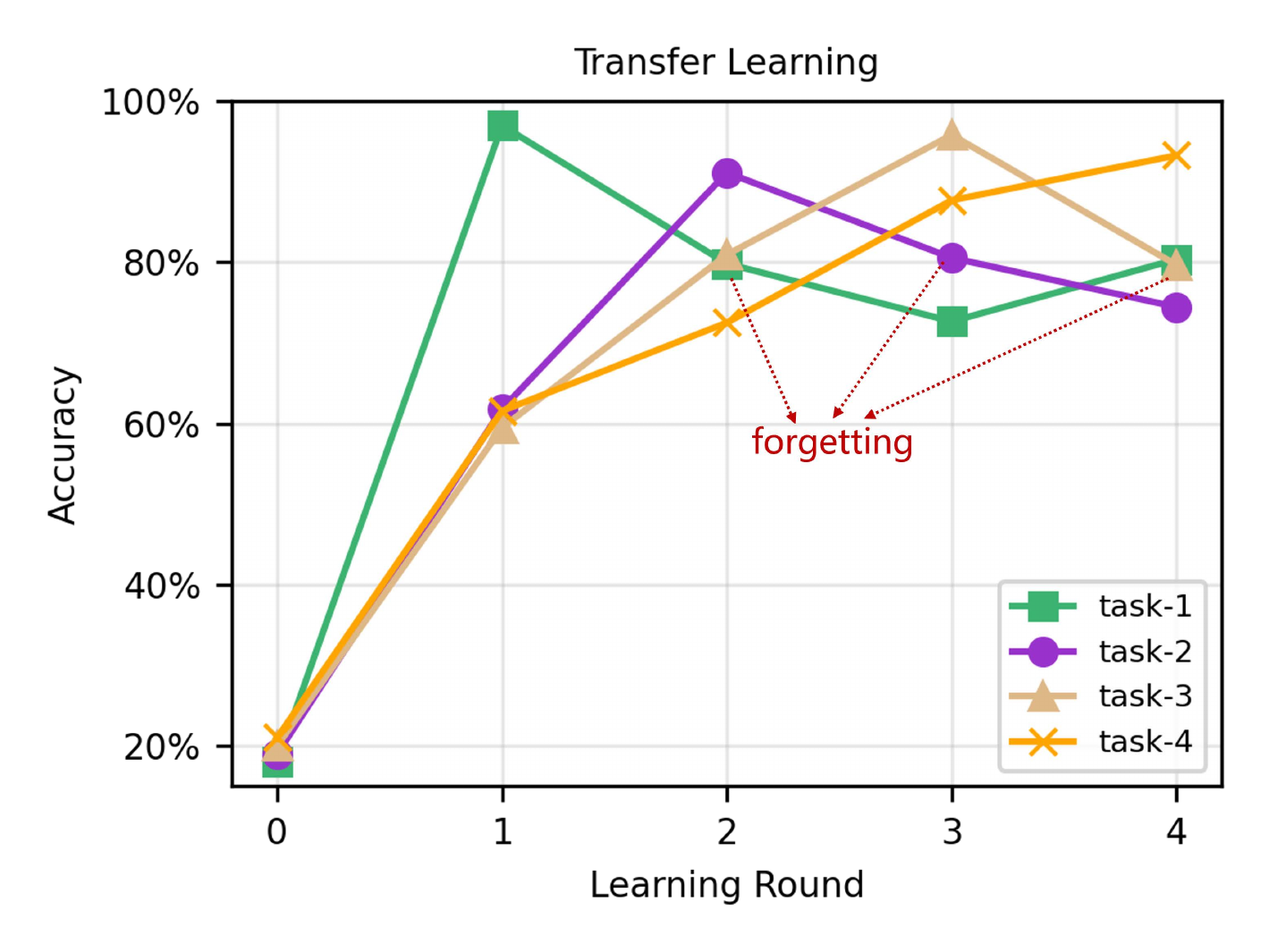}
    \caption{
    When using transfer learning, a model learns four sequential tasks—one in each round. But as the neural network adapts to new tasks, its performance on the first task deteriorates by the second, third, and fourth rounds. This decline stems from weight adjustments for new tasks, which can inadvertently overwrite knowledge from previous tasks.}
    \label{fig2}
\end{figure}

As illustrated in Fig.~\ref{fig1}(a), TL involves adapting a shared backbone network for diverse tasks. However, a significant downside of TL is \textit{catastrophic forgetting}, where the model, while learning new tasks, tends to forget the previously learned knowledge. This phenomenon is demonstrated in Fig.~\ref{fig2}. On the other hand, as depicted in Fig.~\ref{fig1}(b), ML showcases rapid task adaptation, albeit at a computational cost \cite{finn2017model}, which might hinder its application in time-sensitive scenarios.

Given the limitations of TL and ML in addressing data drift, exploring alternative or complementary strategies becomes imperative to mitigate the impact of data drift on mechanical fault diagnosis. Addressing these challenges head-on, we propose a novel approach that seamlessly adapts to dynamic data distributions.
Our system, EdgeFD, leverages a data drift detection module and a weight consolidation mechanisms, allowing for the adaptation and optimization of dynamic data distributions. Our contributions are as follows:



$\bullet$ We introduce a confidence-based data drift detection module that enables real-time monitoring and precise detection of abnormal data drift in bearing FD tasks.

$\bullet$ Our utilization of the continual learning (CL) method based on weight consolidation effectively adapts to drifting data, facilitating incremental learning as new data emerges while efficiently adjusting the drift component.

$\bullet$ For situations with frequent data drift and a high emphasis on long-term learning and knowledge retention, our Drift-Aware Weight Consolidation (DAWC) approach stands out as a superior solution.

$\bullet$ The EdgeFD system harnesses the benefits of edge computing, enabling bearing FD tasks to be conducted on edge devices, minimizing dependence on cloud resources, and data transfer latency, and ensuring real-time responsiveness.

\section{Problem Definition}
\label{section:dawc}
In this section, we outline the problem's core attributes, particularly focusing on the challenges faced when naively implementing solutions based solely on TL and ML.\looseness=-1

\subsection{Problem Definition}
\label{section:problem_definition}
Mechanical equipment often exhibits drift in the underlying data distribution of their operational conditions. Factors like varying revolutions, measured in rpm (revolutions per minute), or different loads, measured in Hp (horsepower), can instigate such drifts.

As mechanical systems function, the FD model is iteratively fine-tuned over a series of tasks, represented as $\{\mathcal{T}^{(1)}, \mathcal{T}^{(2)}, \cdots, \mathcal{T}^{(T)}\}$, where ${\mathcal{T}^{(t)}}$ is a labeled dataset of the $t^{th}$ task, and $\mathcal{T}^{(t)}=\{\boldsymbol{x}_i, \boldsymbol{y}_i\}_{i=1}^{N_t}$, which consists of $N_t$ pairs of instances $\boldsymbol{x}_{i}$ and their corresponding labels $\boldsymbol{y}_i$. Assuming the most realistic situation, we consider the case where the task sequence is a task stream with an unknown arriving order, such that the model can access $\mathcal{T}^{(t)}$ only at the training period of this task $\mathcal{T}^{(t)}$, which will become inaccessible afterward. Given $\mathcal{T}^{(t)}$ and the model learned so far, the learning objective at $t$ is as follows: \[\underset{\bm{\theta}^{(t)}}{\mathop{\min}}\ \mathcal{L}(\bm{\theta}^{(t)}; \bm{\theta}^{(t-1)}, \mathcal{T}^{(t)}),\]
where $\mathcal{L}(\cdot;\cdot)$ is the loss function that quantifies the difference between predictions made using the current model parameters $\bm{\theta}^{(t)}$ and the ground truth data associated with task $\mathcal{T}^{(t)}$.
The role of $\bm{\theta}^{(t-1)}$ is to provide an initial point for the optimization process, enabling the model to adapt quickly and effectively to the new task $\mathcal{T}^{(t)}$.

To solve the optimization problem described above, various methods have been developed that leverage the knowledge contained in $\bm{\theta}^{(t-1)}$ to enhance the learning process for the new task $\mathcal{T}^{(t)}$. \looseness=-1
Particularly, in the context of diagnosis model training, the techniques based on TL and ML have been mostly explored.

\subsection{Transfer Learning and Meta Learning}
\label{sections:tl_ml}
\textbf{Transfer Learning:}
The goal of TL is to speed up the learning of new tasks by sharing knowledge between different tasks. The optimization goal of TL can be expressed as minimizing empirical risk, for example using a cross-entropy loss function
\begin{equation}
\mathop{\min}_{\bm{\theta}^{(t)}} \sum_{t=1}^{t} \mathbb{E}_{(\boldsymbol{x}_i,\boldsymbol{y}_i)\sim \mathcal{T}^{(t)}}[\ell(\boldsymbol{y}_i,f(\boldsymbol{x}_i;\bm{\theta}^{(t)}))],
\end{equation}
where $f(\cdot;\cdot)$ is the corresponding classifier function.
The model parameter for the $t$-th task is the combination of the backbone network $\bm{u}^{(t)}$ and the classifier $\bm{v}^{(t)}$:
$\bm{\theta}^{(t)}:=(\bm{u}^{(t)} \circ \bm{v}^{(t)})$.
The TL process freezes the backbone network $\bm{u}^{(t)}=\bm{u}^{(t-1)}$ and updates only the parameters of the classification head $\bm{v}^{(t)}$.
$\ell(\boldsymbol{y}_i,f(\boldsymbol{x}_i;\bm{\theta}))$ is the loss function, for measuring the difference of the predicted output $f(\boldsymbol{x}_i;\bm{\theta})$ against true labels $\boldsymbol{y}_i$. \looseness=-1

\textbf{Meta-Learning:} The goal of ML is to learn the updating rules from multiple tasks, optimizing the learning process for subsequent tasks.
A popular algorithm in this domain is Model-Agnostic Meta-Learning (MAML), where the optimization objective is
\begin{equation}
\label{eq:maml}
\min_{\bm{\theta}} \sum_{t=1}^{t} \mathbb{E}_{(D_t^{s}, D_t^{q}) \sim \mathcal{T}^{(t)}} \left[ \ell \left( f_{\bm{\theta}'}(D_t^{s}), D_t^{q} \right) \right],
\end{equation}
where $\bm{\theta}'=\bm{\theta} - \alpha \nabla_{\bm{\theta}} \ell(f_{\bm{\theta}}(D_t^{s}), D_t^{q})$, and $f_{\bm{\theta}}(D_t^{s})$ denotes the model prediction on task $\mathcal{T}^{(t)}$.
In \eqref{eq:maml}, $\bm{\theta}$ represents the global model parameters, $D_t^{s}$ is the support set for the task $\mathcal{T}^{(t)}$ (used to fine-tune the model), $D_t^{q}$ is the query set (used to evaluate the performance), $\alpha$ is the learning rate.

While achieving certain successes, it is noteworthy that the existing TL-based FD methods often suffer from catastrophic forgetting (as shown in Fig.~\ref{fig2}), leading to continuous model adaptation and increased computational burden on edge devices. Meanwhile, ML-based approaches struggle to handle sequential tasks and involve expensive second-order derivative computations. Moreover, these studies assume the learner's awareness of task distribution changes, allowing the computation of a new posterior approximation. However, this assumption diverges from real-world IIoT scenarios encompassing gradual, sudden, and repeated drifts.

Therefore, data drift detection is necessary and important before model adaptation, and the detection of data drift can serve as a valuable cue for model adaptation. For example, the drift intensity helps to reduce the search space and ease the learning process \cite{lesort2021understanding}.
Specifically, by dynamically adjusting the consolidation weights of previous parameters based on the drift detection results, we can effectively address the challenges posed by significant data drifts.

\section{Learning Methodology}
In this section, we introduce a novel framework termed DAWC that effectively detects drift in data streams and employs weight consolidation for model fine-tuning. DAWC addresses the challenges faced by TL and ML methods while also reducing computational costs for edge devices.

\subsection{Data Drift Detection}
\label{section:drift_detection}
\begin{figure}[t]
    \centering
    \includegraphics[width=0.8\linewidth]{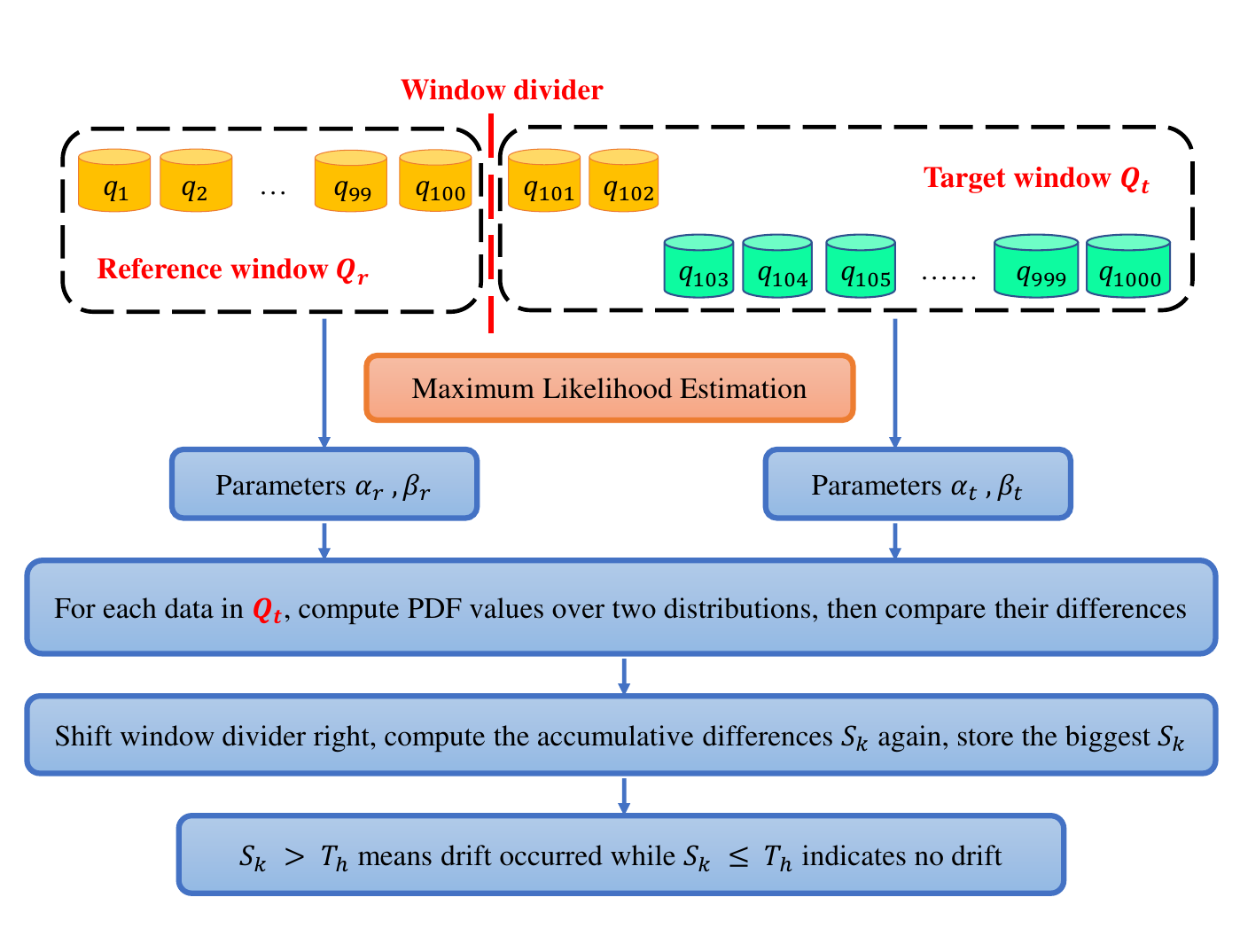}
    \caption{The simplified flow of drift detection algorithm. $T_h$ is the prefixed threshold, which is related to the sensitivity $\lambda$.}
    \label{fig9}
\end{figure}

Inspired by the CUSUM-type \cite{haque2016sand} detection method, we propose a sliding window data drift detection method based on the Beta distribution, as illustrated in Fig.~\ref{fig9}.
In the realm of statistical modeling, the Beta distribution stands out when dealing with random variables bounded within a fixed interval, especially those representing proportions or probabilities. This distribution is naturally defined between 0 and 1, making it a prime choice for situations where we model uncertainties of confidence scores. A key strength of the Beta distribution lies in its versatility; its shape can adapt from U-shaped to bell-shaped, contingent upon its parameters. In our context, the confidence scores from model predictions fall between 0 and 1. As data drift manifests, these confidence score distributions might shift. By leveraging the adaptability of the Beta distribution, we can effectively capture and compare these shifts, making it a powerful tool for drift detection in our methodology.

Within a sliding window of length $N$, the tuple $[\hat{y}_{i},q_i]$ denotes the model's prediction category and the corresponding confidence score for the input data sample $\boldsymbol{x}_i$.
The sliding window is partitioned into two segments: the reference window $Q_r$, housing historical confidences, and the target window $Q_t$, which contains the latest confidences.  It is assumed that these confidence scores in the two windows follow different Beta distributions.
When data drift occurs, the CNN model's classification accuracy declines, causing small confidence values to appear in the target window, leading to discrepancies between the two Beta distributions.
To detect data drift, we compare the cumulative probability density function (PDF) values of the confidences in the target window derived from these two Beta distributions.

\textbf{Step 1}. Estimate Beta Distribution Parameters: Using the Maximum Likelihood Estimation (MLE) method, we estimate the parameters of the Beta distribution. Next, the PDF of the beta distribution is provided by

\begin{equation}
    f(q_i|\alpha,\beta) =
    \begin{cases}
        \frac{q_i^{\alpha-1}(1-q_i)^{\beta-1}}{B(\alpha,\beta)}, & \text{if } 0 < q_i < 1, \\
        0, & \text{otherwise}.
    \end{cases}
\end{equation}

where,
\begin{equation}
B(\alpha,\beta) = \int_{0}^{1}q_i^{\alpha-1}(1-q_i)^{\beta-1}dq_i.
\end{equation}

\textbf{Step 2}. Calculate Dissimilarity Score: Compute the dissimilarity score for PDF values on the two Beta distributions. Sum these scores to obtain the cumulative dissimilarity score $s_k$.

\textbf{Step 3}. Detect Drift: If the cumulative dissimilarity score $s_k$ surpasses a predefined threshold $T_h$, data drift is confirmed.
The prefixed threshold \(T_h\) is a critical value used to determine the occurrence of data drift. Its value is inversely related to sensitivity \(\lambda\), i.e., a smaller \(\lambda\) results in a higher threshold, making the drift detection more stringent.

\textbf{Step 4}. Determine Drift Location: Resize the sub-windows and shift the divider, recalculating the dissimilarity scores. If the new cumulative dissimilarity score $s_f^{'}$ is larger, it implies the new divider is closer to the drift point. This process continues until the maximum dissimilarity score $s_f$ is reached, pinpointing the position of the data drift occurrence.

Algorithm~\ref{alg:detection} provides a structured procedure to detect data drift using the cumulative dissimilarity score. It iteratively evaluates each data point in the sliding window against a threshold to determine the onset of drift.
Given a sliding window $Q$, sensitivity to change $\lambda$, padding $\delta$, and a maximum window size $N_{max}$, the algorithm determines whether a drift has occurred by iterating through the sliding window and evaluating the cumulative dissimilarity score against a threshold. 

\begin{algorithm}[t]
\caption{Drift Detection Algorithm}
\label{alg:detection}
  \SetKwInput{Input}{Input}
  \SetKwInput{Output}{Output}
  \SetKwInput{Parameter}{Parameter}
  \SetKwFunction{InitializeModel}{initialize\_model}
  \SetKwFunction{EstimateParams}{estimateParams}
  \SetKwFunction{Max}{max}
  \SetKwFunction{TestForDrift}{test\_for\_drift}
  \SetKwFunction{Mean}{mean}
  \SetKwFunction{ExtractConfidenceScores}{extract\_confidence\_scores}
  \SetKwFunction{DriftDetection}{drift\_detection}
  \SetKwFunction{FineTuneTransferLearning}{fine\_tune\_transfer\_learning}
  \SetKwFunction{FineTuneMetaLearning}{fine\_tune\_meta\_learning}
  \SetKwFunction{FineTuneDawc}{fine\_tune\_dawc}
  \SetKwFunction{EvaluateModel}{evaluate\_model\_on\_task}
  \SetKwFunction{SampleForDriftDetection}{sample\_for\_drift\_detection}
  \SetKwFunction{Main}{main}

  \Input{
    Sliding window $Q$, Sensitivity to change $\lambda$, Padding $\delta$, Maximum size for the sliding window $N_{max}$.
  }
  \Output{
    The boolean value indicates whether drift is detected.
  }

  \BlankLine

  $s_f \leftarrow 0$,
  $T_h \leftarrow -$log$(\lambda)$,
  $N \leftarrow |Q|$\\

  \BlankLine

  \For{$k \leftarrow \delta$ \textbf{to} $N - \delta$}{
    $m_r \leftarrow \Mean(q_1 : q_k \in Q)$\;
    $m_t \leftarrow \Mean(q_{k+1} : q_N \in Q)$\;

    \BlankLine

    \If{$m_t \leq (1 - \lambda) \cdot m_r $}{
      $s_k \leftarrow 0$\\
      $[\hat{\alpha}_r,\hat{\beta}_r] \leftarrow \EstimateParams(q_1 \colon q_k)$;\\
      $[\hat{\alpha}_t,\hat{\beta}_t] \leftarrow \EstimateParams(q_{k+1} \colon q_N)$;\\

      \For{$i \leftarrow k+1$ \textbf{to} $N$}{
            $s_k \leftarrow s_k + \text{log} \left( \frac{f(q_i|\hat{\alpha}_t,\hat{\beta}_t)}{f(q_i|\hat{\alpha}_r,\hat{\beta}_r)}
            \right)$;\\
      }
      $s_f \leftarrow \text{Max}(s_f,s_k)$;\\
    }
 }

\BlankLine

\If{$s_f > T_h$}{\textbf{return} $true$}
\Else{\textbf{return} $false$}

\end{algorithm}

\subsection{Weight Consolidation}
\label{section:weight_consolidation}
Once the drift is identified using our approach, the next challenge is fine-tuning the model without compromising or forgetting the previously acquired knowledge. Zhang \textit{et al.} \cite{zhang2023spatial} presented a CL method centered on data prototype replay to address this. While effective, this method comes with the downside of increased storage and computational costs. On the other hand, regularization, a tool traditionally employed to mitigate model overfitting, has garnered attention in the CL domain. Within this sphere, the chief avenues for regularization stem from parameter importance estimates \cite{kirkpatrick2017overcoming,aljundi2018memory} and knowledge distillation\cite{buzzega2020dark}.

Building upon these insights, we introduce our approach.
Drawing from the Elastic Weight Consolidation (EWC) \cite{kirkpatrick2017overcoming} methodology, we employ an approximate Bayesian CL strategy, aiming to seamlessly adapt models in continuous learning scenarios.
Let $\bm{\theta}$ be the parameter vector, and consider the posterior of $\bm{\theta}$:
\begin{equation}
\label{eq:bayesian}
    \begin{split}
        p(\bm{\theta}|\mathcal{T}^{(1:t)}) &\propto p(\bm{\theta})\prod_{i=1}^{t}p(\mathcal{T}^{(i)}|\bm{\theta}),\\
        &\propto p(\bm{\theta}|\mathcal{T}^{(1:t-1)})p(\mathcal{T}^{(t)}|\bm{\theta}).\\
    \end{split}
\end{equation}
Here, the factorization in \eqref{eq:bayesian} arises due to the conditional independence assumption of task data.
However, computing the exact posteriors is challenging, leading to Laplace’s approximation:
\begin{equation}
    p(\mathcal{T}^{(t)}|\bm{\theta}) \approx \mathcal{N}(\bm{\theta}_t^*,\mathcal{F}_t^{-1}),
\end{equation}
where mean $\bm{\theta}_t^*$ centered at the maximum a posteriori parameter when learning task $t$, and the precision given by the Fisher information matrix $\mathcal{F}_t$ evaluated at $\bm{\theta}_t^*$.This matrix approximates the Hessian of the negative log-likelihood, ensuring positive semidefiniteness.

The overall optimization objective is to minimize the empirical risk while simultaneously accounting for the significance of parameters from prior tasks through a regularization term:
\begin{equation}
\label{eq:loss}
    -\log p(\mathcal{T}^{(t)}|\bm{\theta}) + \frac{\lambda}{2}\sum_{i}^{t-1}\mathcal{F}_{i}(\bm{\theta}-\bm{\theta}_{i}^*)^2,\\
\end{equation}
where $\lambda$ is a hyperparameter that controls the penalty on important parameters from previous tasks. The term $\mathcal{F}_{i}$ is the importance-weighted regularization term for the previous task's $\mathcal{T}^{(t)}$ parameter. $\mathcal{T}^{(t)}$ represents the current task's data. This regularization mechanism effectively preserves prior knowledge while adapting to new tasks.

More detailed mathematical derivation of Eqs.~\eqref{eq:bayesian}-\eqref{eq:loss} and additional experimental validations will be included in the extended version of this paper, which will provide an in-depth exploration of the presented methodology.

\section{Experiments}
\label{section:experiment}
In this section, we present the experimental settings and evaluate the performances of our method by comparing them with baseline methods. Moreover, we conduct extensive ablation studies to provide a deeper understanding of our method.

\textbf{Baseline methods.}
The chosen baseline methods represent common strategies in the domain of task-oriented learning:
\begin{itemize}
    \item \textbf{STL}: Single-task learning, where each task is treated independently. This serves as the most basic comparison.
    \item \textbf{FCB} \cite{han2019learning}: A naive TL approach that preserves the feature extraction capabilities of the backbone network while fine-tuning only the classification head.
    \item \textbf{MAML} \cite{finn2017model}: An advanced meta-learning technique aiming at finding a model initialization conducive for rapid adaptation to new tasks.
\end{itemize}

\subsection{Datasets and experimental details}
\textbf{Experimental Setup:}
All experiments are conducted on a system equipped with an NVIDIA RTX 3090 GPU, using PyTorch version 1.9.

\textbf{Dataset:}
This study employs the publicly accessible CWRU Bearing Dataset \cite{smith2015rolling} encompassing 10 classification tasks.
These tasks correspond to different failure sites (Inner Race, Outer Race, Rolling Body) and sizes (0.007', 0.014', 0.021'). The dataset encompasses bearing data with failures under diverse loads (0 hp, 1 hp, 2 hp, 3 hp), reflecting variations in real-world data collection conditions. As a result, we segment the drifting task based on these distinct load conditions.

\textbf{Backbone Network:}
Inspired by \cite{zhang2017new}, we adopt the WDCNN as our core network, recognized for its distinct attributes: (1) a broad initial layer featuring convolutional kernels sized $64 \times 1$; (2) stacking multiple compact $3 \times 1$ convolutional kernels following the wide kernels. The wide initial convolutional kernels capture broad features, while the stacked smaller kernels delve into finer details.

\textbf{Training Setting:}
From the total of 5000 data samples, each load condition has 1250 samples. We employ an 80-20 split strategy: 1000 samples from each load condition were used for model training, while the remaining 250 samples (per condition) serve as the test set to evaluate model performance.

\textbf{Performance Metric:}
We use two metrics from \cite{mirzadeh2020understanding} to evaluate algorithms when the number of tasks is large, i.e., Average Accuracy (AA) and Average Forgetting (AF).

\subsection{Main Results}
To ensure the stability and reliability of our results, we conduct three independent repetitions of the experiments. The outcomes of these repeated trials showed some standard deviation, with precise values provided in parentheses. \looseness=-1

Within our experiments, we focus on investigating four distinct heterogeneous/drifting tasks. After completing training on 4000 samples, we aggregate the experimental outcomes as presented in Table~\ref{tab:AA}. Encouragingly, our DAWC showcase a remarkable performance. Specifically, DAWC demonstrates a notable 4.64\% increase in accuracy compared to the current SOTA baseline method (i.e., MAML). It's worth noting that DAWC not only excelled in terms of performance but also exhibited significant advantages in terms of forgetting rate. This underscores the validation of the efficacy of our proposed weight consolidation strategy.

\begin{table}[htb]
\caption{Performance on CWRU dataset}
\label{tab:AA}
\centering
\setlength{\tabcolsep}{2mm}{
\begin{tabular}{cccc}
\toprule
\textbf{Method} & \textbf{Type}      & \textbf{AA(\%)}                                                  & \textbf{AF}               \\ \midrule
STL             & Baseline           & 81.25 \tiny($\pm$ 0.63)                           & 0.22 \tiny($\pm$ 0.02) \\ \midrule
FCB             & Transfer Learning  & 86.67 \tiny($\pm$ 0.72)                           & 0.16 \tiny($\pm$ 0.02) \\
MAML            & Meta-Learning      & 89.28 \tiny($\pm$ 0.16)                           & 0.14 \tiny($\pm$ 0.01) \\ \midrule
DAWC(ours)      & Continual Learning & \textbf{93.92 \tiny($\pm$ 0.15)} & 0.07 \tiny($\pm$ 0.01) \\ \bottomrule
\end{tabular}
}
\end{table}

\begin{figure}[htb]
    \centering
    \includegraphics[width=0.8\linewidth]{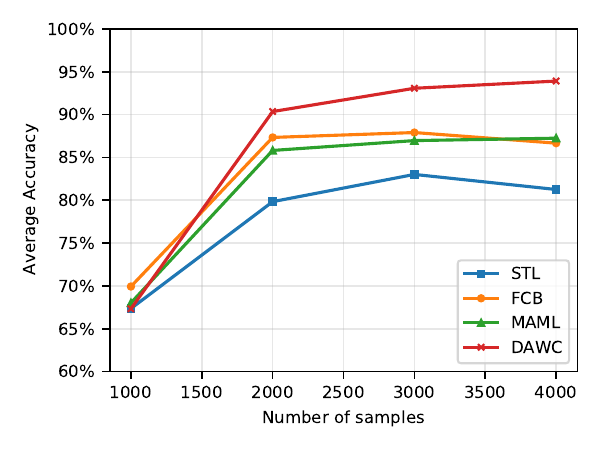}
    \caption{The comparison of the AA over the 4 tasks during the learning phase.}
    \label{fig4}
\end{figure}

Through Fig.~\ref{fig4}, we can observe the trends in AA variations throughout the learning process for the four different methods. Notably, our DAWC approach demonstrates superior performance, notably in the face of multiple instances of data drift. This serves as evidence of the considerable potential of DAWC in handling heterogeneous tasks.

In summary, the DAWC method outperformed the existing baseline methods in terms of both accuracy and forgetting rate, thereby validating the effectiveness of the proposed weight consolidation strategy in handling drifting tasks.

\subsection{Effectiveness of Core Designs}
\begin{figure}[htb]
    \centering
    \includegraphics[width=0.8\linewidth]{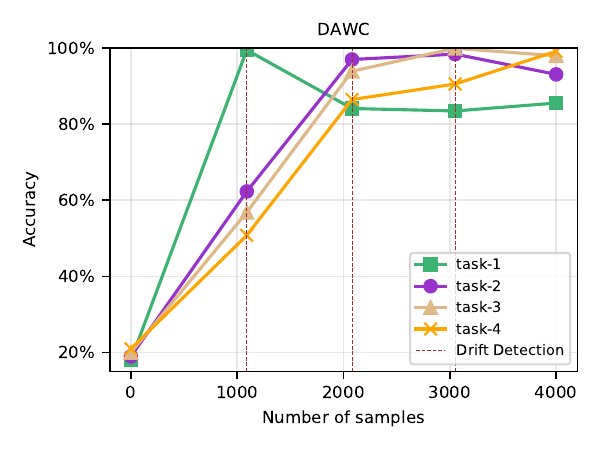}
    \caption{Accuracy for the four tasks in DAWC. The vertical dashed line indicates the time at which a change in the distribution was detected}
    \label{fig10}
\end{figure}

\textbf{Effect of Drift Detection:}
We conduct experiments using the parameters $\lambda=0.05$, $N_{max}=1000$, and $\delta=100$.
Each alteration in bearing load signifies a shift in the fundamental data distribution, indicating data drift. The instances where these shifts occur are 1000, 2000, and 3000.
In Fig.~\ref{fig10}, the vertical dashed lines indicate where a drift is detected by applying our detection method. We can notice that all drifts are detected shortly after they theoretically occur.

\begin{figure*}[htb]
    \centering
    \includegraphics[width=0.9\linewidth]{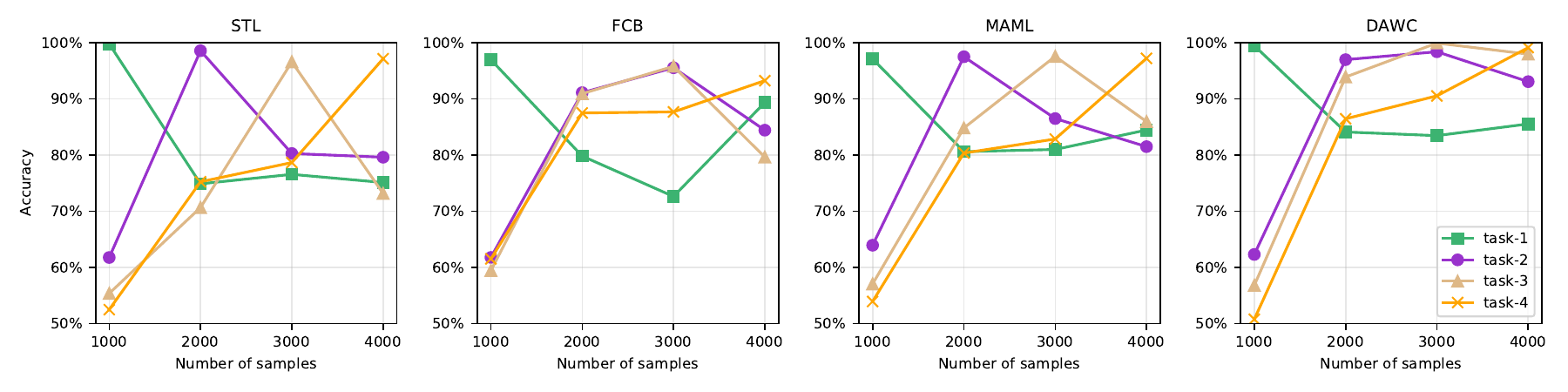}
    \caption{Accuracy comparison of four methods across sequential tasks.}
    \label{fig5}
\end{figure*}

\textbf{Effect of Weight Consolidation:}
We evaluate the accuracy of the four methods for each task, as depicted in Fig.~\ref{fig5}.
It is observed that the accuracy steadily improves with fewer fluctuations as the training for new tasks progresses, highlighting the robustness of our method.

The effectiveness of our approach in mitigating forgetting can be attributed to efficient weight consolidation, which helps preserve previously learned information.

\begin{figure}[htb]
    \centering
    \includegraphics[width=0.8\linewidth]{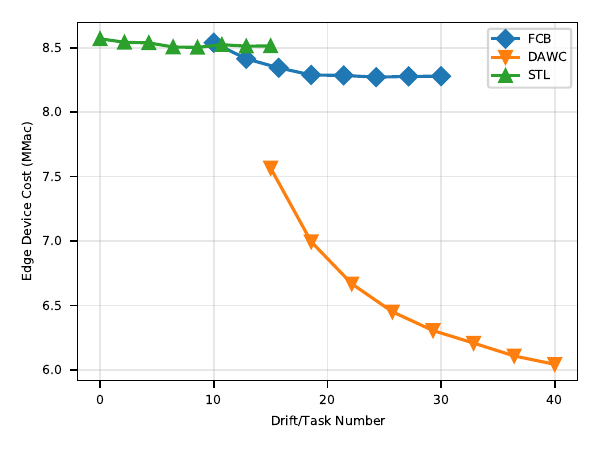}
    \caption{Comparison of average computational costs for drift adaptation.}
    \label{fig6}
\end{figure}

\textbf{Computational Costs:}
Fig.~\ref{fig6} illustrates the accuracy variations for each task throughout the learning process for the three methods. When confronted with multiple instances of data drift, both TL and DAWC demonstrate a certain reduction in computational costs for edge devices compared to the STL method. However, our approach outperforms TL in this aspect.

This superiority arises from the continuous triggering of model fine-tuning thresholds in TL, where parameters are adjusted to accommodate data drift. In contrast, our DAWC scheme rapidly adapts to various data drift scenarios without the need to trigger model fine-tuning thresholds. As a result, it significantly alleviates computational expenses, and it is hence edge-friendly.

\section{System Demo}
\label{section:system}
\begin{figure}[htb]
    \centering
    \includegraphics[width=0.8\linewidth]{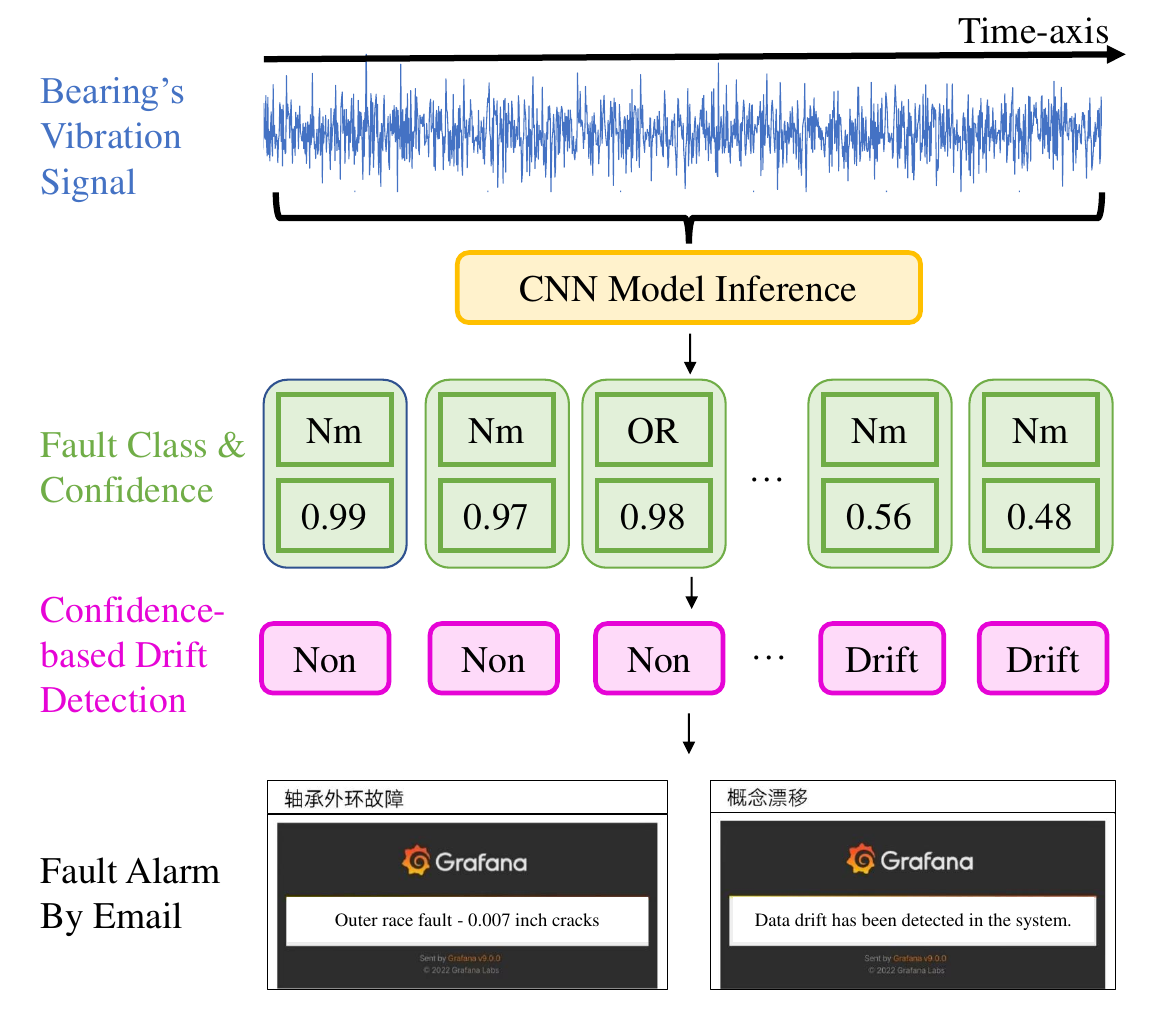}
    \caption{Illustration of the system's operational workflow.}
    \label{fig3}
\end{figure}

Based on the framework of DAWC, we develop a system for bearing FD and alerting. The operational workflow of the entire system is depicted in Fig.~\ref{fig3}, showcasing the major components and stages involved. This system leverages the power of DAWC to discern and classify 10 different types of bearing faults, returning corresponding confidence scores. Upon the detection of a bearing fault, the system promptly sends fault alert notifications to administrators, allowing for the timely initiation of necessary maintenance measures.

\begin{figure}[htb]
    \centering
    \includegraphics[width=0.8\linewidth]{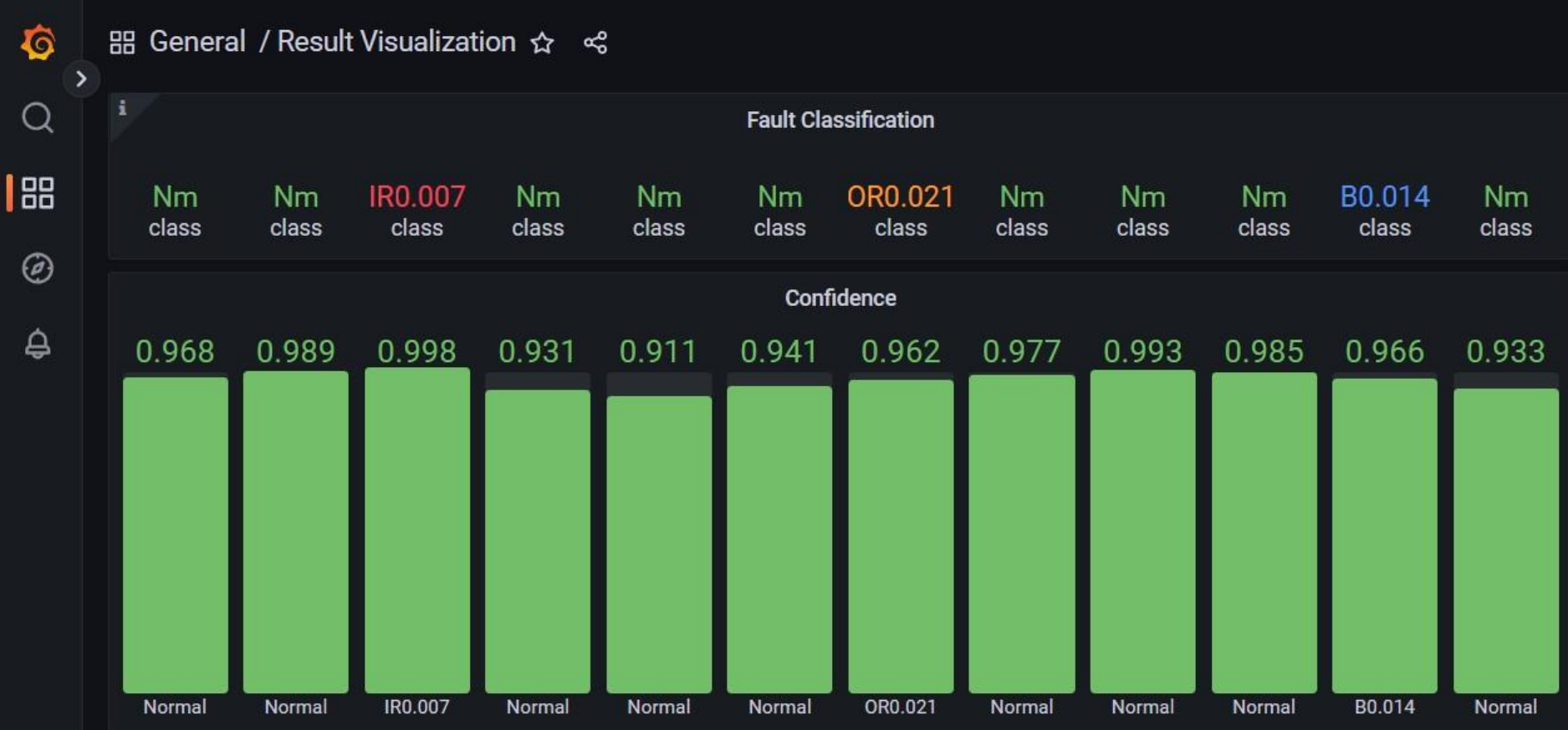}
    \caption{Visualization of the system interface, displaying various fault classifications and their corresponding confidence scores.}
    \label{fig7}
\end{figure}

To provide users with a comprehensive understanding of the system's status and diagnostic results, we design an intuitive visualization interface, as partially depicted in Fig.~\ref{fig7}. At the top of the interface, distinct identifiers for various fault categories are presented. Specifically, ``Nm" indicates ``Normal", ``IR0.007" stands for ``0.007-inch Inner Race Fault", ``OR0.021" represents ``0.021-inch Outer Race Fault", and ``B0.014" means ``0.014-inch Ball Fault". Additionally, beneath each fault classification, the upper section of a rectangular bar displays the associated confidence score, allowing users to grasp the system's confidence in each fault classification.

\begin{figure}[htb]
    \centering
    \includegraphics[width=0.8\linewidth]{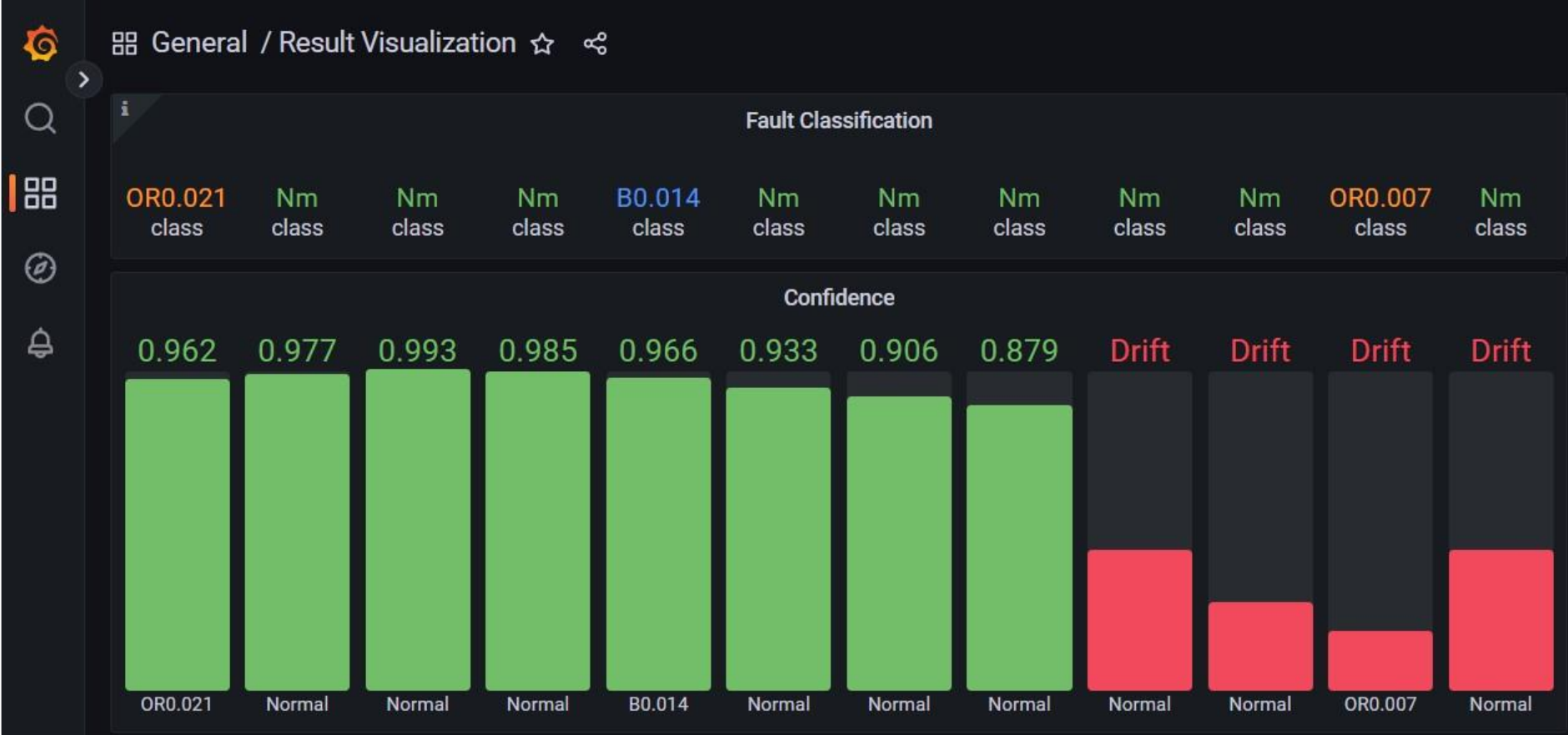}
    \caption{Highlighting the system's approach to drift data, using the term ``Drift" and a red bar for low-confidence drift instances.}
    \label{fig8}
\end{figure}

Next, delving into the system's approach towards handling drift data, as showcased in Fig.~\ref{fig8}, when the system's confidence in drift data is comparatively low, it employs the term ``Drift" accompanied by a red bar, ensuring that users can promptly recognize instances of data drift.

Through our meticulous design, we not only achieve significant progress in bearing FD and alerting but also provide users with the convenience of seamlessly monitoring and understanding the system's operational status.

\balance
\section{Conclusion}
\label{section:conclusion}
In this study, we have presented an edge-oriented method for mechanical fault diagnosis that incorporates drift detection and weight consolidation mechanisms. Distinctively, our method stands out for its significant reduction in computational overhead, especially in scenarios plagued by frequent data drift, offering a resource-efficient solution for edge devices. A prominent characteristic of this method is its simplicity, making it adaptable to various backbone network architectures. This attribute paves the way for leveraging more powerful pre-trained models in future large-scale edge-cloud collaborative settings. By combining this approach with advanced pre-trained models, we can further enhance the performance and scalability of fault diagnosis systems.

\bibliographystyle{IEEEtran}
\bibliography{IEEEabrv,references}
\end{document}